# Plastic Waste Classification Using Deep Learning: Insights from the WaDaBa Dataset

Suman Kunwar [1]
Faculty of Computer Science,
Selinus University of Sciences and Literature,
Ragusa, Italy
Email: sumn2u@gmail.com

Banji Raphael Owabumoye
Department of Geography,
Obafemi Awolowo University,
Ile-Ife, Nigeria
Email: owabumoye@gmail.com

Abayomi Simeon Alade
Department of Physics,
University of Ibadan,
Oyo, Nigeria
Email: abayomy.alade@gmail.com

## Abstract

With the increasing use of plastic, the challenges associated with managing plastic waste have become more challenging, emphasizing the need of effective solutions for classification and recycling. This study explores the potential of deep learning, focusing on convolutional neural networks (CNNs) and object detection models like YOLO (You Only Look Once), to tackle this issue using the WaDaBa dataset. The study shows that YOLO-11m achieved highest accuracy (98.03%) and mAP50 (0.990), with YOLO-11n performing similarly but highest mAP50(0.992). Lightweight models like YOLO-10n trained faster but with lower accuracy, whereas MobileNet V2 showed impressive performance (97.12% accuracy) but fell short in object detection. Our study highlights the potential of deep learning models in transforming how we classify plastic waste, with YOLO models proving to be the most effective. By balancing accuracy and computational efficiency, these models can help to create scalable, impactful solutions in waste management and recycling.

**Keywords:** Plastic Waste Classification; Deep Learning; YOLO Models; WaDaBa Dataset; Sustainable Waste Management

## 1. Introduction

Plastics have become indispensable in modern society due to their versatility, durability, and low cost. As a result, global plastic production has skyrocketed to over 450 million tons annually (Ren et al., 2024). However, the very properties that make plastics so useful, durable, and resilient also contribute to their persistence in the environment, where they can take hundreds of years to degrade. The United Nations Environment Programme (UNEP, 2020) estimates that "more than 8 million tons of plastic waste enter the oceans annually," leading to catastrophic effects on marine life and ecosystems. The most effective strategies to address the global issue of plastic pollution remain uncertain. Borrelle et al. (2020) and Lau et al. (2020) have explored potential solutions and their implications, concluding that substantial reductions in plastic waste generation are achievable over the next few decades if immediate and robust action is taken. However, even under the best circumstances, large amounts of plastic are still expected to accumulate in the environment.

Despite increased awareness and efforts to improve waste management, only a small fraction of plastic waste is effectively recycled. According to recent studies, less than 9% of plastic produced globally is recycled, underscoring inefficiencies in existing waste management frameworks (Jambeck et al., 2015; Geyer et al., 2017). Traditional waste management systems, reliant on manual sorting, are not efficient enough to handle the growing volume of plastic waste. To address this issue, the Society of the Plastics Industry introduced Resin Identification Codes (RICs) in 1988. These codes categorize plastics based on polymer content, facilitating recycling and waste management. Common RICs include PET, PE-HD, PVC, PE-LD, PP, and PS, which are often labeled on plastic products to aid in sorting (Ren et al., 2024).

Artificial intelligence (AI) and machine learning (ML) have emerged as transformative technologies in automating waste management and introduced various techniques to effectively manage waste (Kunwar et al., 2024; Abdu et al., 2023). The use of deep learning models such as YOLO (You Only Look Once) has shown significant promise in improving waste classification accuracy (Kumar et al., 2023; Vimal Kumar et al., 2024). As noted in recent studies, "YOLO's ability to detect objects in real-time with high accuracy makes it ideal for waste sorting applications" (Chen et al., 2021). Furthermore, transfer learning models like MobileNetV2 and RestNet have demonstrated efficiency in recognizing complex waste categories, making them suitable for deployment in resource-constrained environments (Shetty, 2021; Kunwar, 2024).

 These advancements signal a shift towards AI-driven solutions that hold the potential to significantly enhance the efficacy of plastic waste classification and recycling efforts.The aim of this study is to find the effectiveness of the YOLO models in identifying plastic waste using the WaDaBa dataset and compare it with MobileNetV2, Resnet50 and Efficient Net along with custom model in various settings. We also implemented various data augmentation techniques to improve dataset diversity and model generalization. Later, the effective model embedded in mobile app.

## 2. Related Works

Plastic waste management has been a focus of recent research, especially given the limitations of traditional sorting techniques. According to Chen et al. (2021), "traditional waste management approaches are increasingly inadequate given the complexities of modern waste streams."

Studies have shown that integrating AI and deep learning into waste management can significantly improve the efficiency of sorting processes (Olawade et al., 2024).The WaDaBa dataset, which contains 4,000 images categorized into five RIC classes, has become a benchmark for testing various deep learning models (Ren et al., 2024). Traditional methods such as histogram analysis have been used in the past but struggled to generalize across different plastic types. A recent study demonstrated that "deep learning models, when combined with data augmentation, can achieve much higher classification accuracies compared to conventional approaches" (Shetty, 2021).

YOLO models have emerged as the preferred choice for real-time object detection due to their efficiency and accuracy. According to Redmon and Farhadi (2018), "the architecture of YOLO allows for simultaneous object detection and classification, making it faster than traditional CNN-based approaches." One of the studies shows that the use of the larger YOLO v5 model outperforms the smaller, nano and medium sized model on detecting plastic waste detection along railway lines (Liu et al., 2022). The latest version, YOLO-11n, incorporates advanced modules to enhance detection capabilities, particularly for small objects, thus making it highly effective for waste classification (Ren et al., 2024).

In addition to YOLO, transfer learning models like MobileNet V2 have demonstrated high classification accuracy while maintaining computational efficiency. Studies have shown that "models pre-trained on large datasets like ImageNet can significantly improve performance in specialized tasks such as waste detection" (Chollet, 2017).

## 3. Methodology

This study used a comprehensive methodology encompassing data preparation, model training, and performance evaluation. The models included YOLO models, a custom deep learning model, and pre-trained transfer learning models. Each model underwent specific preprocessing and data augmentation to improve classification accuracy.

### 3.1 Dataset and Preprocessing

The WaDaBa dataset comprises 4,000 high-resolution RGB images (1920x1277 pixels, 300 dpi) representing five distinct RIC categories: PET, PE-HD, PP, PS, and Others. These images were captured under varying lighting conditions and angles, with key attributes such as plastic type, color, and deformation levels embedded in the filenames. However, the dataset exhibits a significant class imbalance, with PET accounting for 55% of the total images, while the "Others" category represents just 1%. To address this imbalance, data oversampling was employed to expand the dataset to 11,000 images. Additionally, data augmentation techniques, including random zoom, rotation, contrast adjustment, and flipping, were applied to improve the model's robustness and ability to generalize across different conditions. The distribution of the original dataset is illustrated in Figure 1.

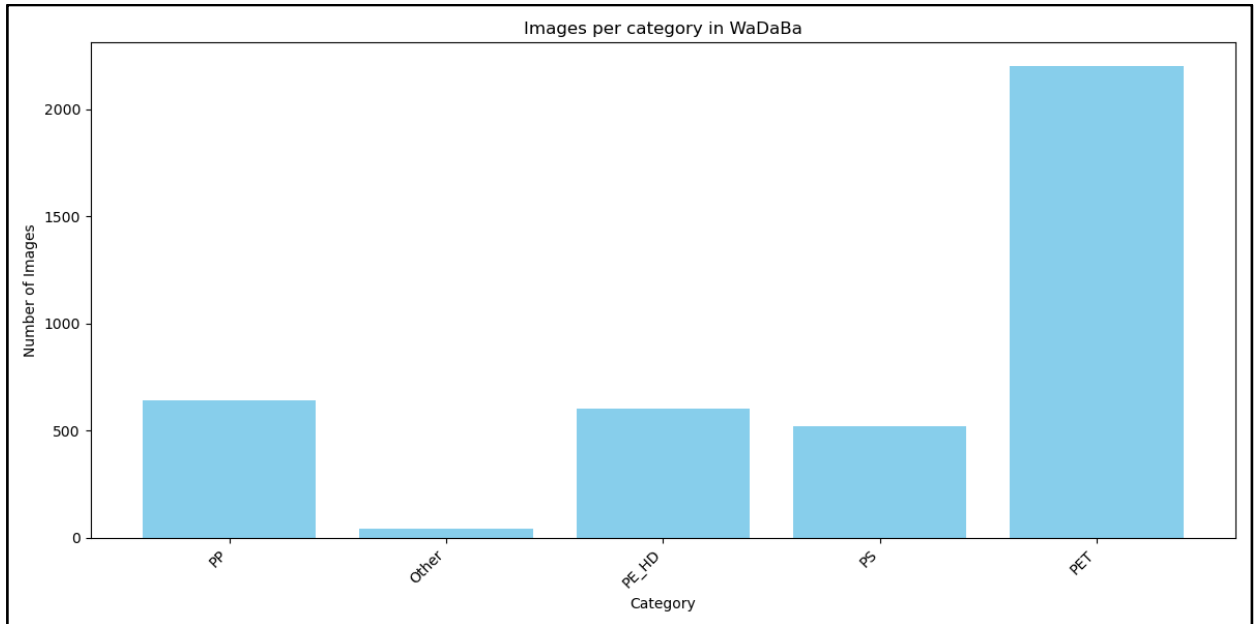

Figure 1. Distribution of the dataset

These images were annotated using annotate-lab, converted into YOLO format and then splited into train and test folders using 80:20 ratio. Figure 2 shows the annotation of an image using annotate-lab (Kunwar, 2024).

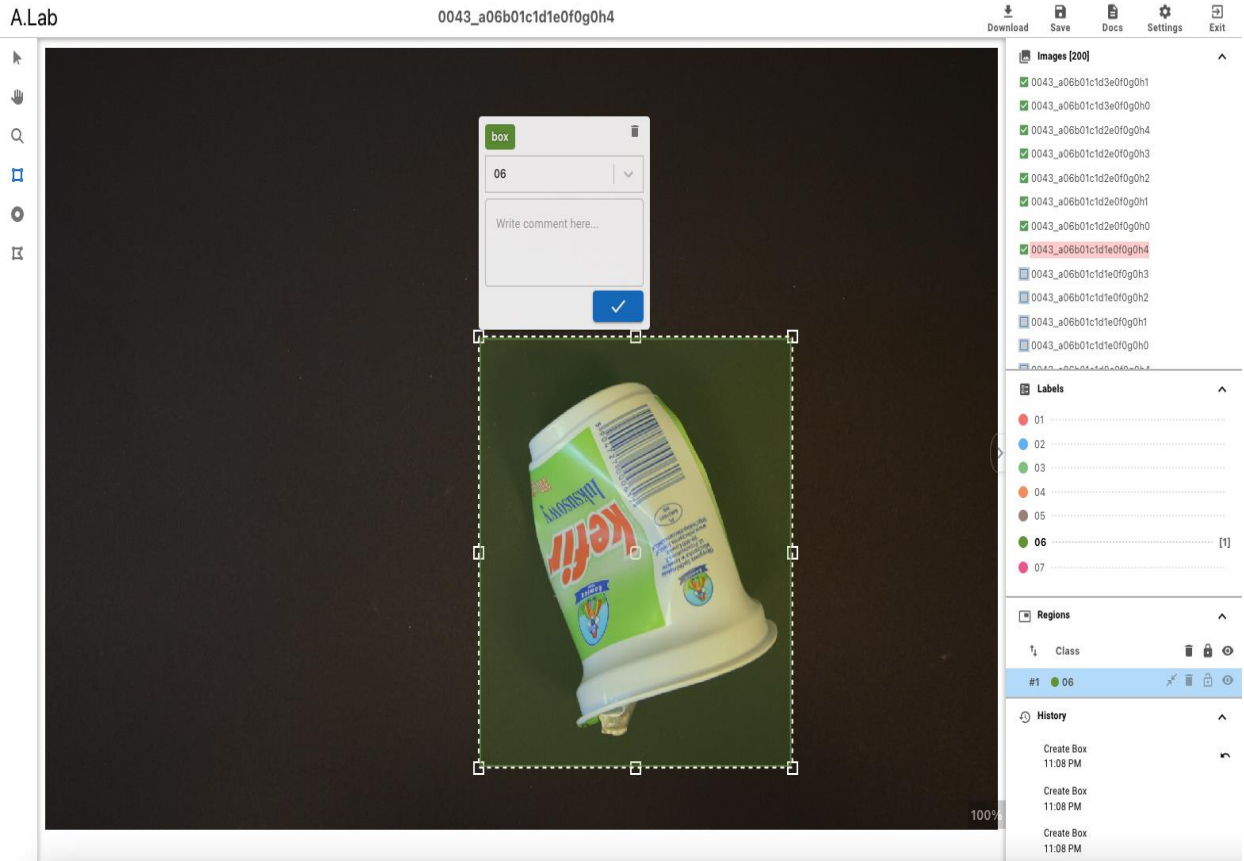

Figure 2. Annotation of WaBaDa image using annotate-lab

## 3.2 Model Architecture and Training

### i. YOLO Model Training

The YOLO models employed in this study are YOLO-10n, YOLO-10m, YOLO-11n, and YOLO-11m. These models were chosen for their exceptional real-time detection capabilities, making them highly suitable for efficient waste classification tasks. These models were meticulously configured to achieve an optimal balance between precision, recall, and mean Average Precision (mAP), which are critical for accurate identification of plastic waste types.

Each YOLO model was rigorously trained over 20 epochs, leveraging non-max suppression techniques to eliminate overlapping bounding boxes and enhance detection confidence. This approach not only improves the precision of detected objects but also significantly reduces false positives, thereby increasing the models' reliability in real-world applications.

The workflow commenced with careful annotation and preparation of the dataset, ensuring compatibility with the YOLO framework. The annotated data were then processed and fed into each YOLO model for training. By fine-tuning hyperparameters, each model was optimized to extract meaningful features, enabling it to accurately classify diverse types of plastic waste under varying conditions.

Following model evaluation, the best-performing YOLO configuration was selected based on its mAP, accuracy, precision, recall, and F1-score metrics. This top-performing model was then further refined through quantization and optimization to enhance its speed and efficiency, making it suitable for deployment in real-time waste management systems. The process flow, which is illustrated in Figure 3, highlights the seamless integration of data preparation, model training, and deployment, emphasizing the robustness of the YOLO models in automated plastic waste detection.

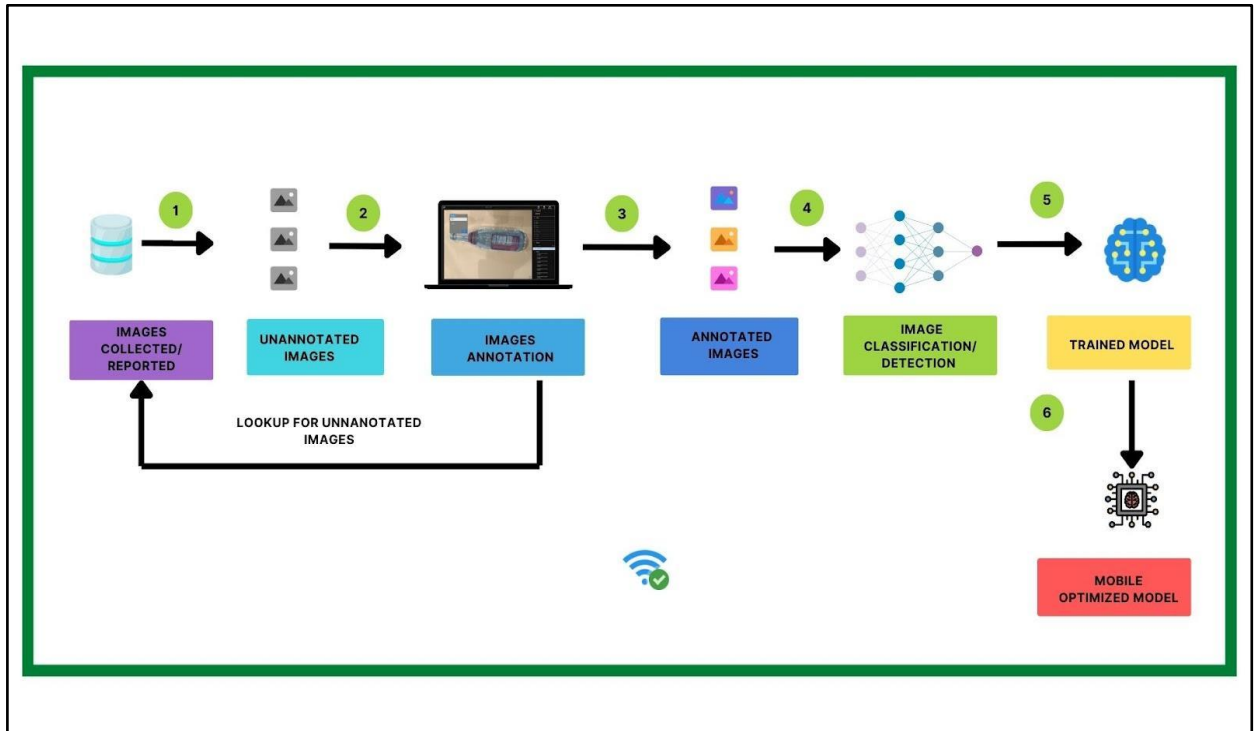

Figure 3. YOLO process flow diagram.

The train and validation loss during model training of YOLO-10N and YOLO-11N is shown in Figure 4 and can see the overall training and validation box and class loss are more in the same pane of smoothness for YOLO-11N with relative to YOLO-10N.

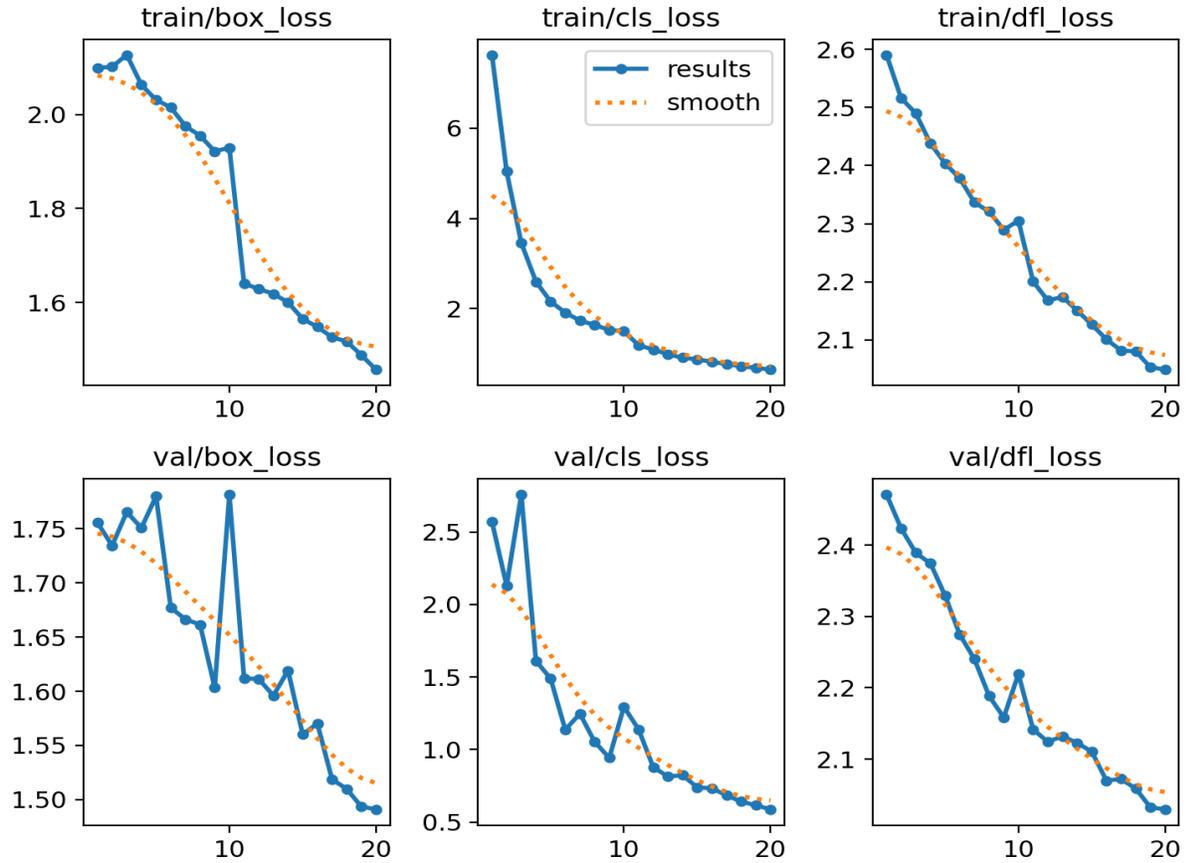

A) Training YOLO-10N

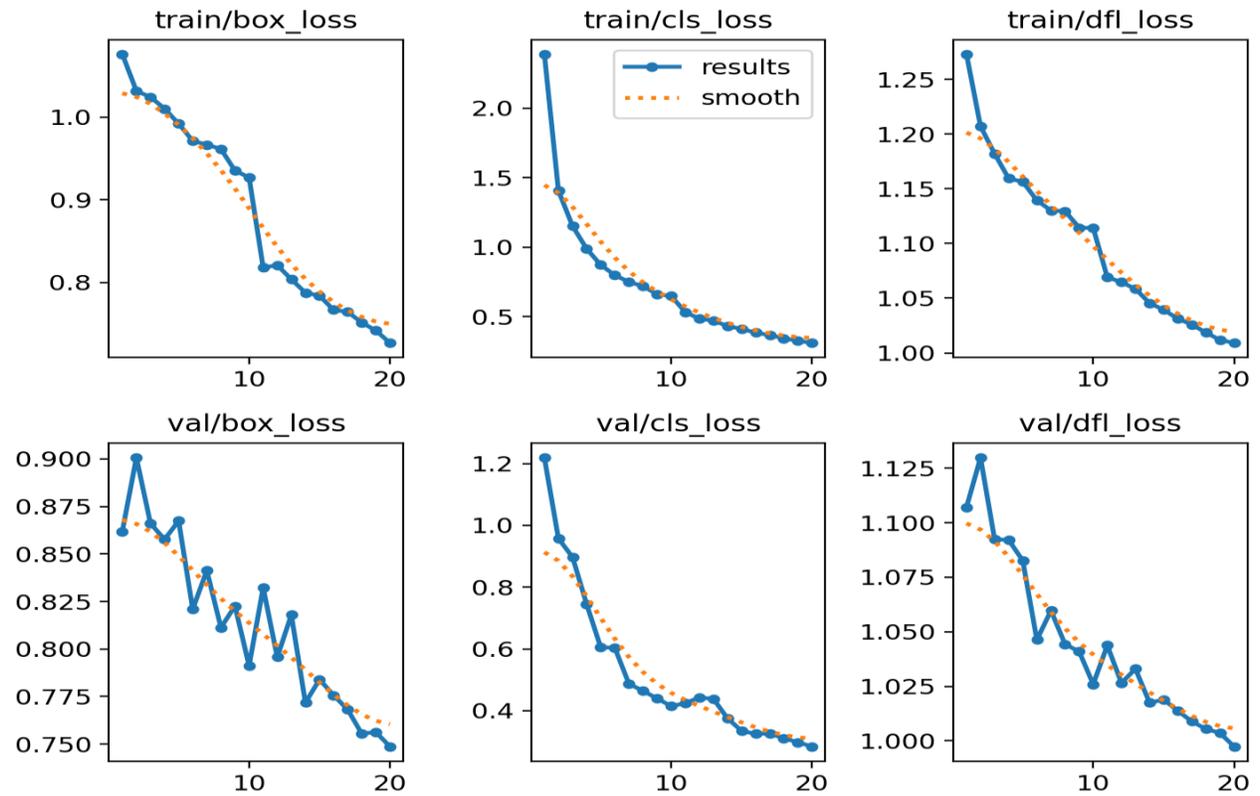

B) Training of YOLO-11N

Figure 4. Training of YOLO-10N and YOLO-11N

## ii. Custom Model Design and Training

The custom model developed in this study was specifically designed to address the challenge of class imbalance present in the dataset. The architecture was thoughtfully constructed to maximize feature extraction and ensure robust classification accuracy.

The model's architecture included a series of three convolutional layers, utilizing filter sizes of 16, 32, and 64, respectively. Each convolutional layer was paired with max-pooling and ReLU activation functions to efficiently capture intricate patterns in the data. To prevent overfitting, a dropout rate of 50% was incorporated across two dense layers containing 256 and 64 neurons, enhancing the model's generalization capability.

For training, the model was configured to run over 40 epochs, utilizing an 80-20 split for training and testing. To optimize computational efficiency while maintaining accuracy, all images were resized to 180 x 180 pixels. A batch size of 300 was employed, leveraging the computational power of two GPUs to expedite training. The final output layer utilized SoftMax activation to effectively handle the multi-class classification task.

The architecture of the model, along with its configuration details, is illustrated in Figure 5, highlighting the seamless integration of feature extraction, dimensionality reduction, and classification components. This carefully crafted architecture demonstrates the model's capability to adapt to the diverse characteristics of the dataset while maintaining high classification performance.

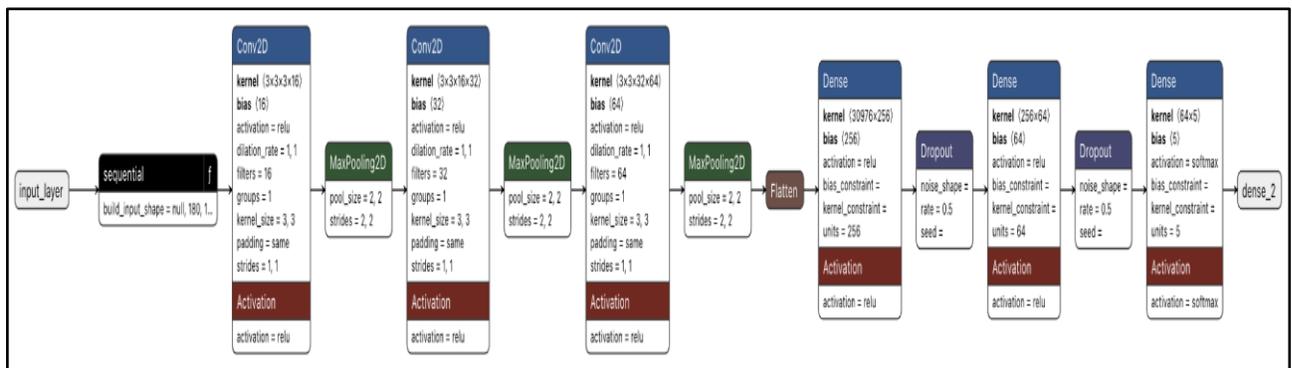

Figure 5. Custom model architecture.

Figure 6 illustrates loss and validation loss and accuracy and validation accuracy during training over epochs and shows MobileNetV2 and Custom model with high accuracy with less loss compared to others.

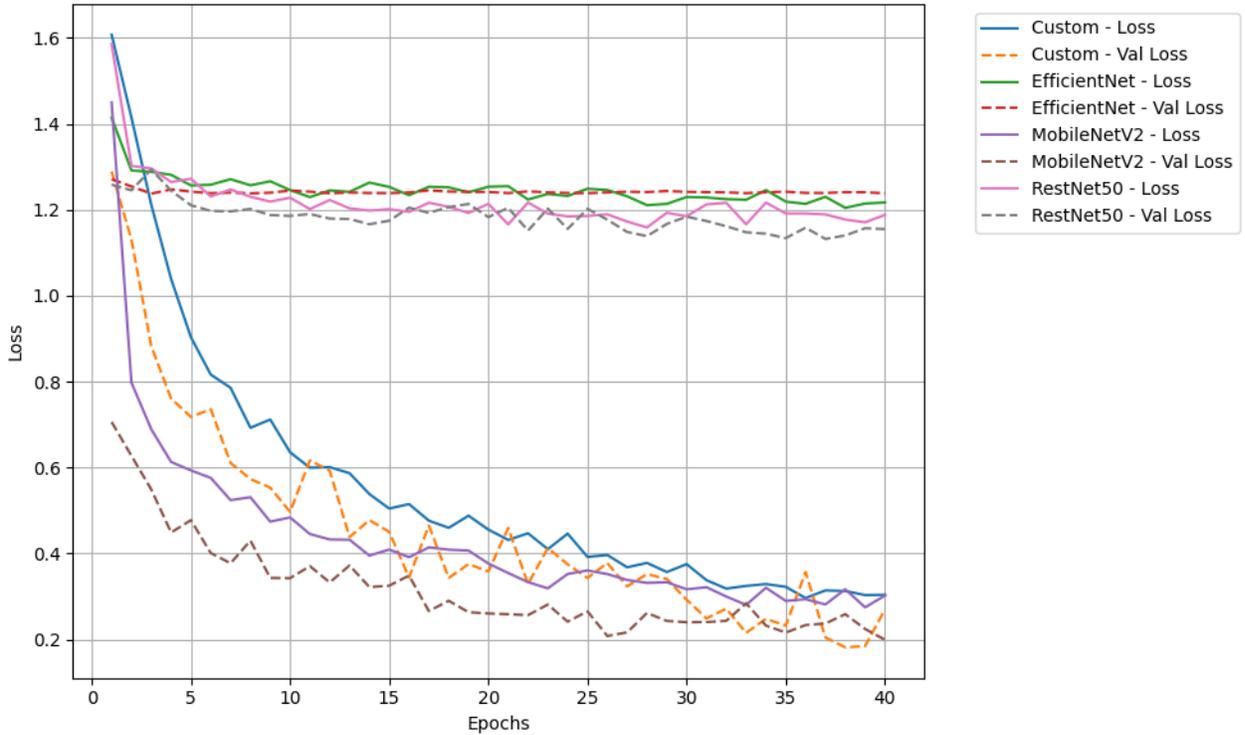

A) Loss and Validation Loss vs Epochs for different models

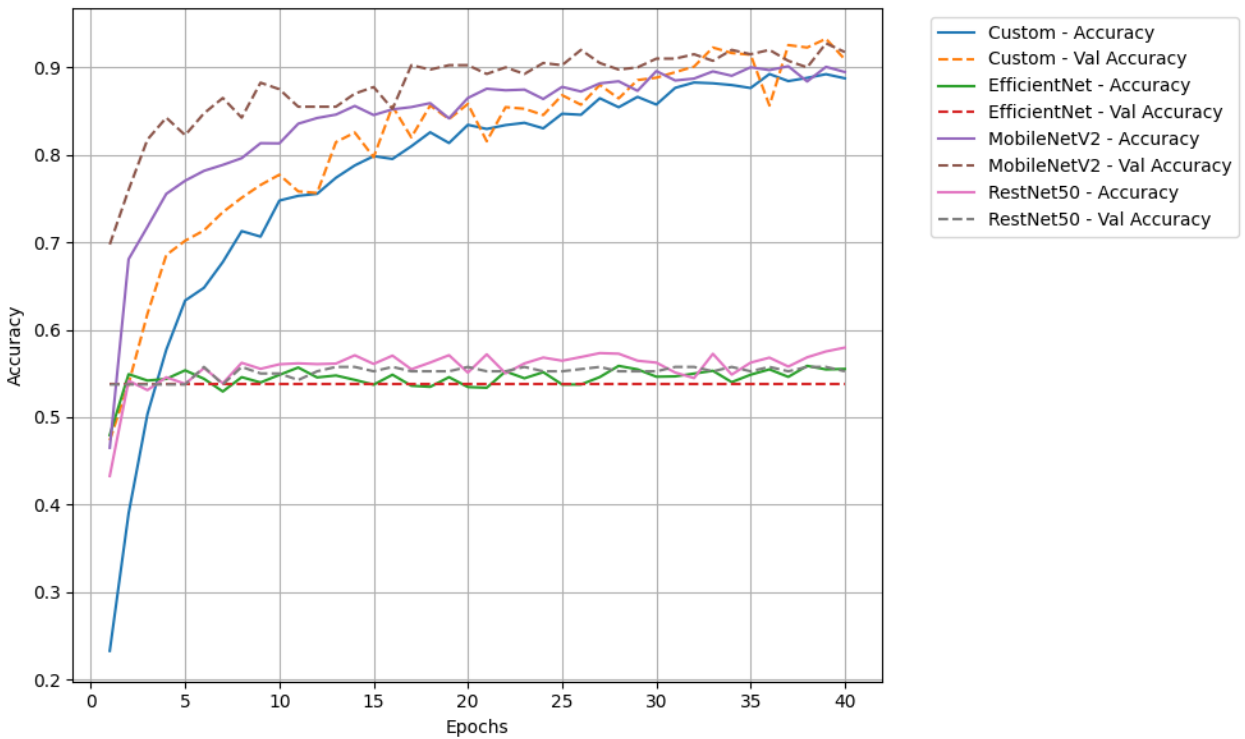

B) Accuracy and Validation Accuracy vs Epochs for different models

Figure 6. Accuracy, Loss, Validation Accuracy and Validation Loss at various epochs during model training

Transfer learning was applied to MobileNet V2, ResNet50, and EfficientNet, which were pre-trained on ImageNet. The following configurations were applied:

1. Data Augmentation: Random zoom, contrast, rotation, and flipping techniques were applied to the training images to simulate real-world variations.
2. Training Configuration: A batch size of 100 and a learning rate of 1e-3 were used, with images resized to 224 x 224 pixels. Two hidden layers with 256 and 64 neurons were added, and SoftMax was applied to the output layer for classification.

The transfer learning models evaluated include MobileNet V2, ResNet50, and EfficientNet. These models were fine-tuned on the WaDaBa dataset using a batch size of 100 and learning rate of 1e-3. As noted in the literature, "transfer learning can significantly reduce the training time while improving model accuracy" (Chollet, 2017).

## 4. Results

The models were evaluated using a range of performance metrics, including accuracy, precision, recall, F1-score, and mAP, with each model demonstrating unique strengths across different evaluation criteria.

### a. YOLO Models

The YOLO-10n model exhibited well-rounded performance, achieving a precision of 0.9161, recall of 0.9460, F1-score of 0.9842, and mAP50 of 0.984, making it highly suitable for general applications where a balanced performance across all metrics is required.

The YOLO-10m model, with an emphasis on recall, achieved a recall of 0.9374, an F1-score of 0.9563, and a mAP50 of 0.956, making it ideal for scenarios where prioritizing recall is critical, such as in detection tasks with high importance on identifying true positives. YOLO-11n set a new benchmark for accuracy, attaining the highest accuracy of 0.9553, along with a precision of 0.9803 and a mAP50 of 0.992. These results make YOLO-11n particularly suitable for accuracy-intensive applications that demand precise classifications.

YOLO-11m, focused on minimizing false positives, achieved a precision of 0.9814 and a mAP50-95 of 0.815, making it the optimal choice for applications where reducing false positives is paramount, such as in high-stakes detection environments. The YOLO-11 models demonstrated superior performance, offering an excellent balance between accuracy and computational efficiency, making them ideal for practical deployment across a wide range of applications. The model predictions and their corresponding results are presented in Figure 7.

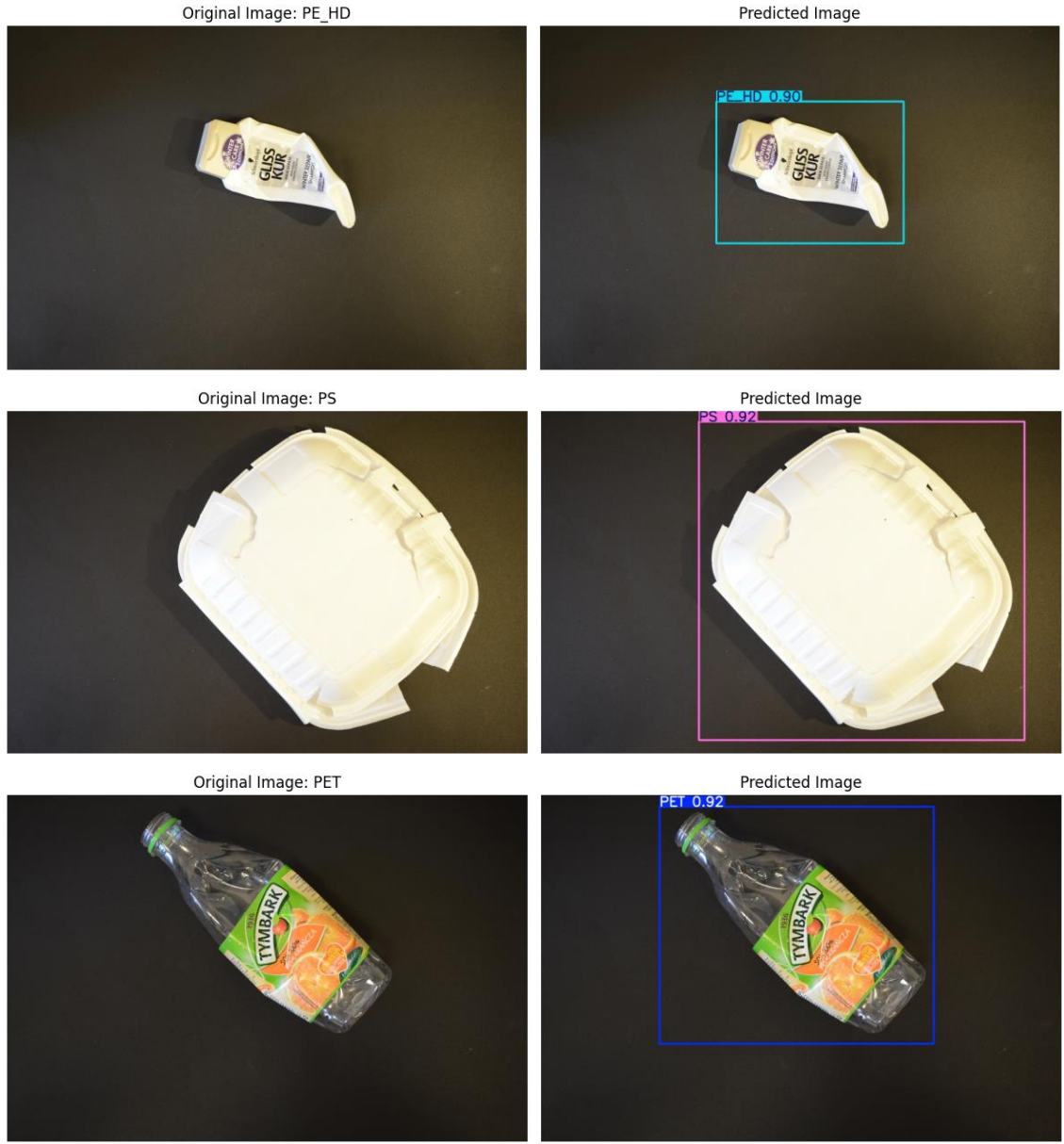

Figure 7. YOLO models' predictions on unseen data.

## b. Custom and Transfer Learning Models

The custom model demonstrated impressive performance, achieving an accuracy of 93.05%, with precision, recall, and F1-score of 92.96%, 93.06%, and 92.98%, respectively. Trained over 40 epochs, the model completed its training in a swift 8 minutes and 1 second. Remarkably, it successfully classified all samples in an unseen test set of 8 images with complete accuracy.

These results highlight the model's robustness and consistency, underscoring its effectiveness in accurately distinguishing between various plastic waste categories. Such high performance across multiple evaluation metrics indicates that the custom model is well-suited for practical waste classification tasks. The outcomes of its predictions on unseen data are depicted in Figure 8, further validating its generalization capability.

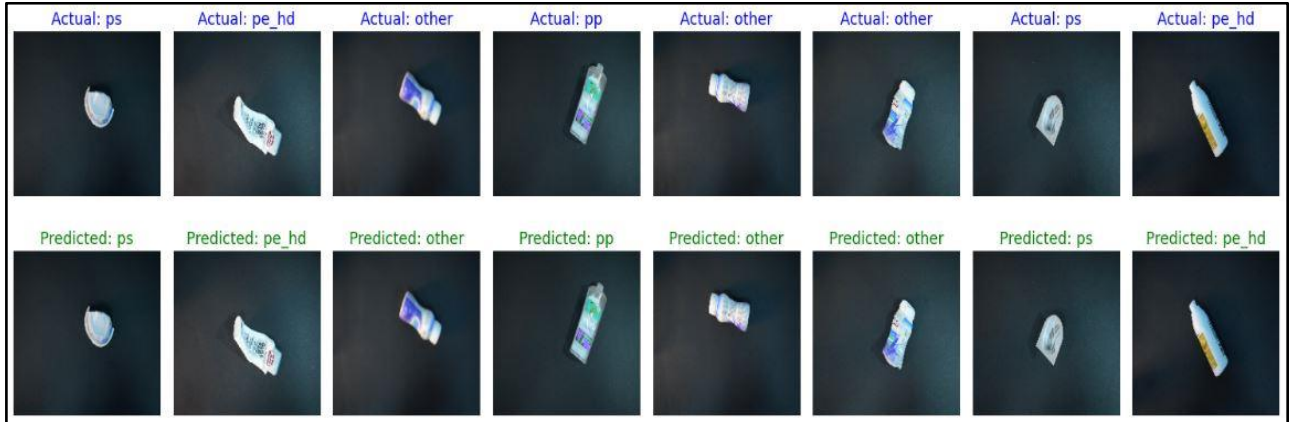

Figure 8. Custom model predictions on unseen data.

MobileNet V2 emerged as the top performer among the transfer learning models, achieving an impressive accuracy of 97.12%. It also demonstrated exceptional precision (96.31%), recall (92.69%), and F1-score (94.26%) after a 40-epoch training period. Notably, MobileNet V2 achieved these results with a remarkably short training time of just 5 minutes and 34 seconds, highlighting both its efficiency and predictive strength. The model successfully classified every image in the sample test set without error, underscoring its robust performance. The outcomes of the model's predictions on unseen data are presented in Figure 9.

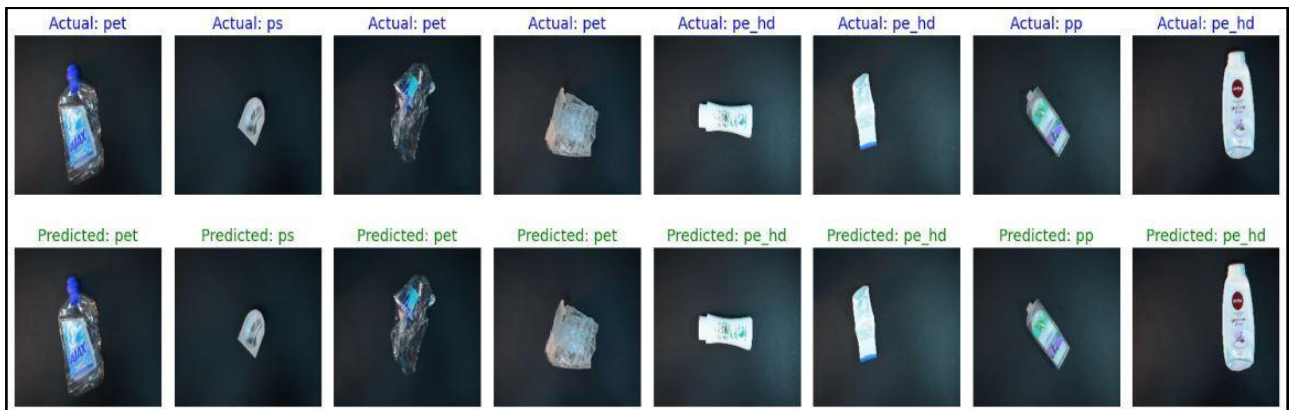

Figure 9. MobileNet V2 predictions on unseen data

ResNet50 exhibited lower performance in this experiment, achieving an accuracy of 65.20%, precision of 65.43%, recall of 45.89%, and an F1-score of 20.01%. Despite a training time of 8 minutes and 31 seconds over 40 epochs, the model correctly classified only 5 out of the 8 samples in the test set. These results suggest that while ResNet50 may demonstrate efficacy in other contexts, it faced challenges with this dataset, due to the pronounced class imbalance. The predictions made by the model on unseen data are illustrated in Figure 10.

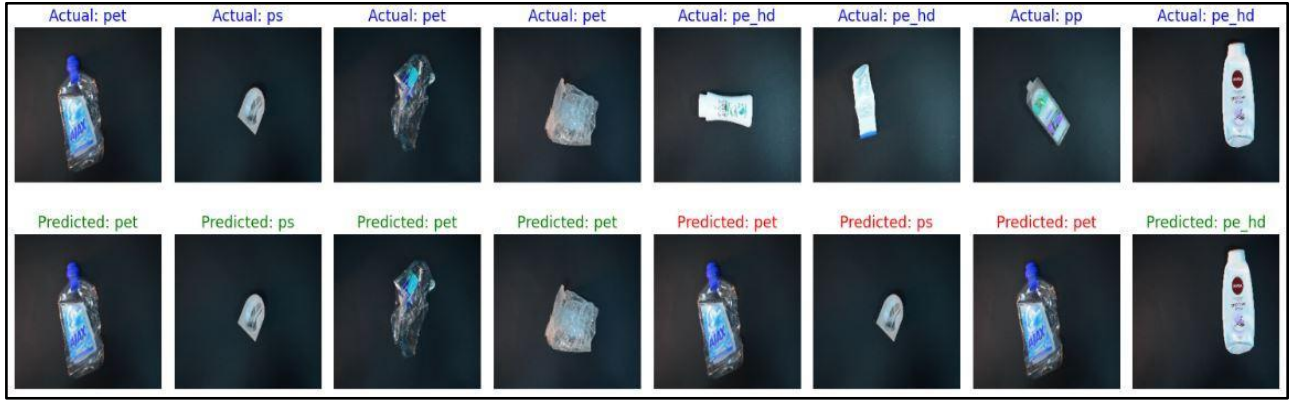

Figure 10. ResNet50 predictions on unseen data

EfficientNet demonstrated limited performance, achieving an accuracy of 53.25%, precision of 10.65%, recall of 20.00%, and an F1-score of 13.90%. With a training duration of 6 minutes and 35 seconds, the model successfully predicted only 3 out of the 8 test images. This relatively poor performance may be attributed to the challenges posed by the dataset's class imbalance, which hindered the model's ability to generalize effectively, particularly for minority classes. The model's predictions on unseen data are shown in Figure 11.

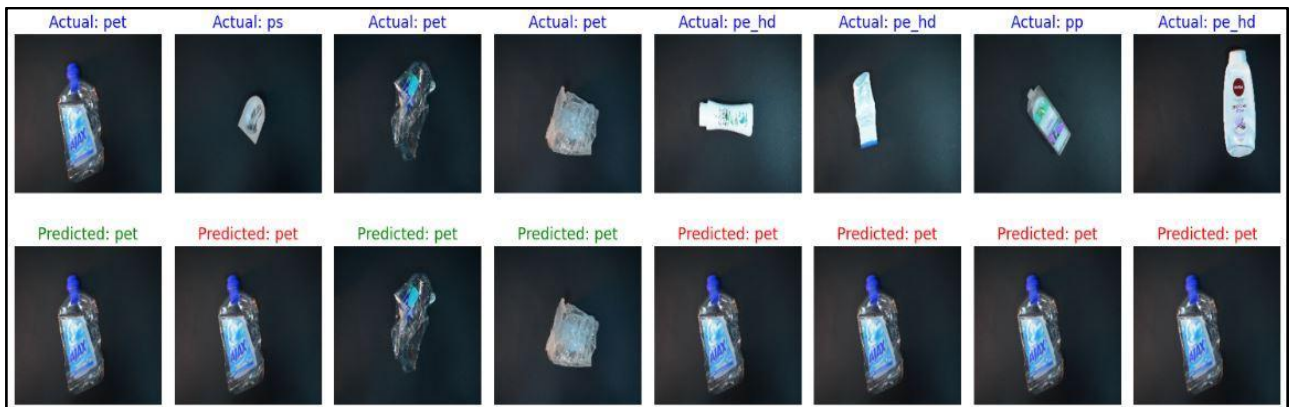

Figure 11. EfficientNet predictions on unseen data

Results indicate that YOLO models, especially YOLO-11n and YOLO-11m, and MobileNet V2 are most effective for plastic waste classification. Table 1 summarizes the performance metrics for each model.

Table 1: Comparison of performance metrics of various models

| Model | Accuracy | Precision | Recall | F1-score | mAP50 | mAP50-95 | Epoch | Training Time |
|-------|----------|-----------|--------|----------|-------|----------|-------|---------------|
| YOLO-10n | 0.9161 | 0.9460 | 0.9842 | 0.8706 | 0.984 | 0.807 | | 27m:50s |
| YOLO-10m | 0.8101 | 0.9374 | 0.9563 | 0.7685 | 0.956 | 0.78 | | 43m:12s |
| YOLO-11n | 0.9803 | **0.9740** | **0.9921** | **0.9553** | **0.992** | 0.813 | 20 | 25m:34s |
| YOLO-11m | **0.9814** | 0.9657 | 0.9908 | 0.9483 | 0.991 | **0.815** | | 42m:40s |
| Custom | 0.9286 | 0.9280 | 0.9285 | 0.9276 | - | - | | 8m:01s |
| MobileNet V2 | 0.9712 | 0.9631 | 0.9269 | 0.9426 | - | - | 40 | 5m:32s |
| RestNet50 | 0.6520 | 0.6543 | 0.4589 | 0.2001 | - | - | | 8m:31s |
| EfficientNet | 0.5325 | 0.1065 | 0.2000 | 0.1390 | - | - | | 6m:35s |

Figure 12 illustrates the confusion matrices of YOLO-11n and YOLO-11m where the YOLO-11n outperformed the YOLO-11m.

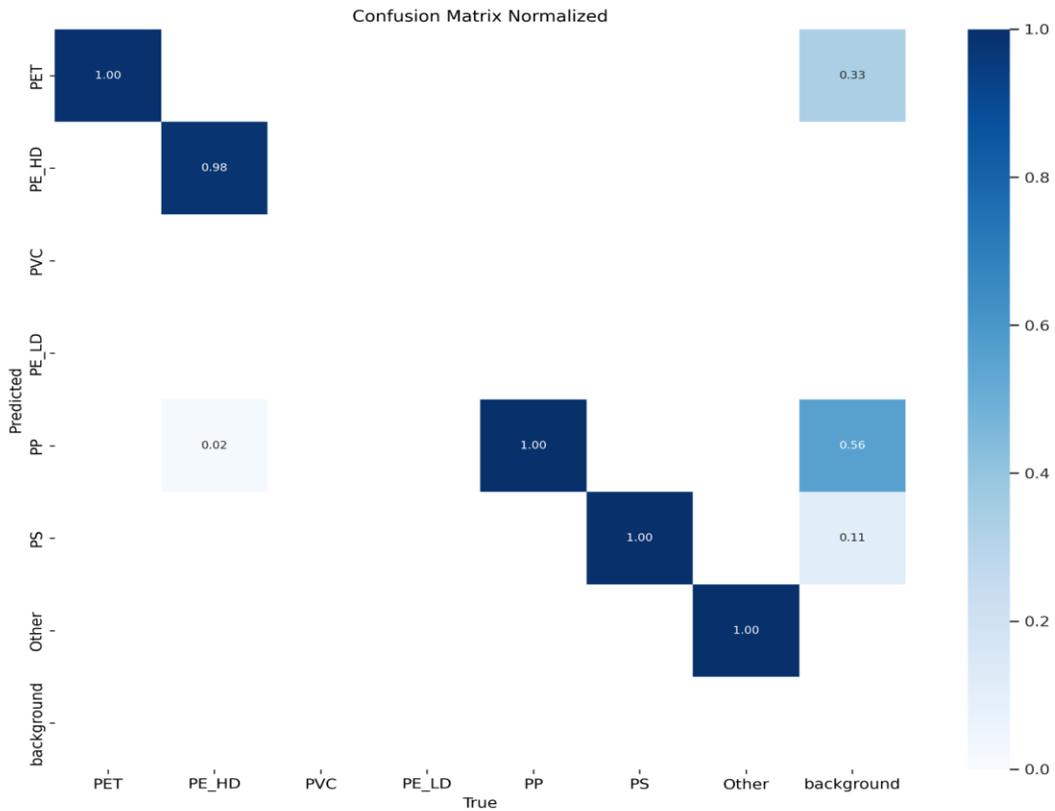

A)  YOLO 11n - Confusion Matrix

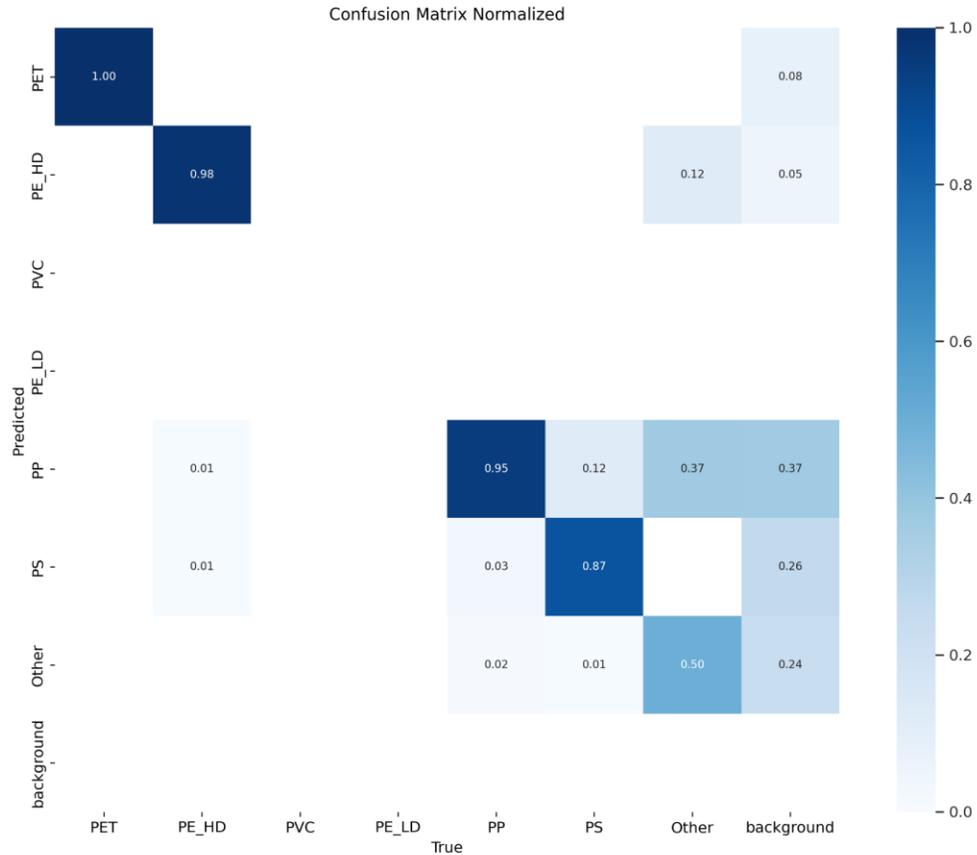

B) YOLO 11m - Confusion Matrix

Figure 12. Confusion Matrix of YOLO-11n and YOLO-11m

## 5. Discussion and Conclusion

In this study, we explored the detection of plastic waste using WaDaBa dataset using YOLO-10n, YOLO-10m, YOLO-10m and YOLO-11n and benchmark it with Custom, MobileNetV2, RestNet50 and EfficentNet  models using transferring learning. Their performances were measured using performance metrics such as accuracy, precision, recall, f1-score, along with training time.  For YOLO models the mAP50 and mAP50-95 were also measured. Our experiments provide insights into the performance of these models, revealing critical aspects that influence their suitability for this domain. The custom model, was designed specifically to handle the imbalanced nature of the WaDaBa dataset, demonstrated strong classification capabilities. By employing targeted data augmentation strategies such as random rotation, zoom, contrast adjustments, and flipping. We significantly improved the diversity of training samples. This enhancement increased the robustness of the custom model and improved its accuracy and generalization. Oversampling techniques were also applied to balance class distribution, which mitigated biases during the training process and further optimized model performance. These findings align with recent studies that emphasize the importance of data augmentation and class balancing for improving model accuracy in datasets with skewed distributions (Ren et al., 2024).

From our experiment we found that MobileNet V2 emerged as the most effective, achieving a classification accuracy of 97%. The lightweight architecture along with the pre-trained ImageNet weights leverage rich feature representations allowing to better generalization to unseen data compared to custom model. Among the pre-trained models, MobileNet V2 emerged as the most effective, achieving a classification accuracy of 97%. Its lightweight architecture, designed for efficient deployment in mobile and edge computing environments, provided a significant advantage in balancing speed and accuracy. The use of pre-trained ImageNet weights enabled MobileNet V2 to leverage rich feature representations, allowing it to generalize better to unseen data compared to the custom model. This aligns with the findings on the efficiency of MobileNet V2 in resource-constrained settings (Shetty, 2021).

The custom model also showed competitive performance, demonstrating the potential of tailored architectures when designed with domain-specific data augmentation. In contrast, deeper architectures like ResNet50 and EfficientNet did not perform as well, with accuracies of 65% and 53%, respectively. These models struggled with the WaDaBa dataset due to its inherent characteristics and the complexity associated with imbalanced data. This observation supports previous research suggesting that deeper models are not always the optimal choice when the dataset is limited in size or requires domain-specific feature extraction (Chollet, 2017). The suboptimal performance of ResNet50 and EfficientNet underscores the importance of choosing models that are appropriately scaled to the dataset's requirements, particularly when computational resources are limited.

Additionally, our findings align with previous work on the application of deep learning for waste classification. For example, Chen et al. (2021) demonstrated the real-time object detection capabilities of YOLO models, which is consistent with the superior performance we observed using the YOLO-11 series in earlier phases of this study. However, our results indicate that in scenarios where computational resources are restricted, leveraging lightweight models like MobileNetV2 or optimized custom architectures may provide a more effective solution than relying on deeper models such as ResNet50 or EfficientNet.

This study highlights the potential of combining custom models with pre-trained, lightweight architectures like MobileNet V2 to improve the accuracy of plastic waste classification. The integration of data augmentation and class balancing techniques demonstrated significant performance improvements particularly for the custom architecture. Future work should explore further refining and generalization of the WaDaBa dataset, experimenting with more advanced augmentation techniques, and developing real-time solutions for automated waste management systems.

**Acknowledgments**

The authors confirm that the data used in this experiment are obtained with permission and can be available on request. The experiments can be downloaded from https://github.com/sumn2u/wadaba-analysis

## Competing Interest

The authors declare no conflict of interest.